\documentclass[letterpaper, 10 pt, conference]{ieeeconf}  

\IEEEoverridecommandlockouts                              

\usepackage{CJKutf8}                                      
\usepackage{verbatim}                                     
\usepackage{graphicx}                                     
\usepackage{amssymb}                                      
\usepackage{mathtools}
\usepackage{multirow}
\usepackage{multicol}
\usepackage{amsfonts}
\usepackage{ctable} 
\usepackage{siunitx}
\usepackage{csquotes} 
 
\graphicspath {{./Fig/}}

\usepackage{graphics} 
\begin{document}

\title{\LARGE \bf
Acceleration of Actor-Critic Deep Reinforcement Learning for Visual Grasping in Clutter by State Representation Learning Based on Disentanglement of a Raw Input Image}

\author{Taewon Kim$^{*}$, Yeseong Park$^{*}$, Youngbin Park and Il Hong Suh%
\thanks{Taewon Kim, Yesung Park, Youngbin Park, and Il Hong Suh are with the Department of Electronics and Computer Engineering, Hanyang University, Korea,
{\tt\small incorl.ktw@incorl.hanyang.ac.kr, ysp1026@incorl.hanyang.ac.kr, pa9301@hanyang.ac.kr, ihsuh@hanyang.ac.kr}}%
\thanks{*Equal contribution}%
}

\maketitle
\thispagestyle{empty}
\pagestyle{empty}

\begin{abstract}

For a robotic grasping task in which diverse unseen target objects exist in a cluttered environment, some deep learning-based methods have achieved state-of-the-art results using visual input directly. In contrast, actor-critic deep reinforcement learning (RL) methods typically perform very poorly when grasping diverse objects, especially when learning from raw images and sparse rewards. To make these RL techniques feasible for vision-based grasping tasks, we employ state representation learning (SRL), where we encode essential information first for subsequent use in RL. However, typical representation learning procedures are unsuitable for extracting pertinent information for learning the grasping skill, because the visual inputs for representation learning, where a robot attempts to grasp a target object in clutter, are extremely complex. We found that preprocessing based on the disentanglement of a raw input image is the key to effectively capturing a compact representation. This enables deep RL to learn robotic grasping skills from highly varied and diverse visual inputs. We demonstrate the effectiveness of this approach with varying levels of disentanglement in a realistic simulated environment.

\end{abstract}


\section{INTRODUCTION}

In this study, we focus on robotic grasping using actor-critic deep reinforcement learning (RL), in which the ability to grasp diverse and previously unseen objects is directly learned from raw image pixels and rarely observed rewards. Recently, deep learning-based robotic grasping methods based on supervised or self-supervised learning techniques \cite{Levine16}, \cite{Mahler18} have achieved state-of-the-art results in tasks where high-dimensional visual data is directly used as input to deep neural networks and diverse unseen target objects are deployed in a cluttered environment. Deep RL has also been applied to a wide range of continuous control tasks in recent years, including robotic grasping \cite{Haarnoja18}, \cite{Lillicrap16a}, \cite{Popov17}. Although deep RL-based continuous control has received significant attention, the performances attained by these methods are not as surprising as those produced by supervised or self-supervised learning methods.





The Deep Q-Network (DQN) \cite{Mnih15} provides a simple and practical scheme for RL with image observations and is one of the most popular methods in value-based and off-policy RL classes. However, incorporating continuous actions, such as joint angles in grasping applications, induces a massive increase in computation. DQNs must perform additional optimization every time it selects a continuous action. This prevents DQNs from being practical in continuous control. 

In contrast, off-policy actor-critic algorithms such as the deep deterministic policy gradient (DDPG) \cite{Lillicrap16a} and distributed distributional deep deterministic policy gradient (D4PG) \cite{Barth18} algorithms are by nature applicable to continuous control without massive computational cost. This is because the actor, which predicts an action at any given state via a single feed-forward computation, is explicitly trained in this class of RL algorithms. However, these methods performs very poorly when grasping diverse objects, especially when learning from raw images and with sparse rewards \cite{Quillen18}. According to \cite{Quillen18}, learning an action-value function and its corresponding actor function jointly makes them co-dependent on each other's output distribution. Thus, learning can eventually become unstable and difficult to tune. In this context, it could be valuable to investigate how vision-based grasping skills for diverse objects in clutter can be learned using off-policy actor-critic deep RL methods such as DDPG and D4PG.  

Most deep RL techniques learn a state representation as a part of learning values or policies via typical RL objectives. This might make learning the vision-based grasping skill difficult, especially in an actor-critic deep RL setting where there is co-dependency between the actor and critic. To make these types of RL techniques feasible for visual grasping tasks, we adapt state representation learning (SRL) techniques. We first encode essential information using SRL to then use it for RL. Although explicitly learning a compressed latent representation to improve performance in RL is not a novel concept, we have not found many studies that have applied this idea to explore the diverse and highly varied situations that arise in the robotic grasping scenarios considered in this study. In particular, we found that the typical SRL procedure is unsuitable for extracting pertinent information for the subsequent learning of the grasping skill performed by deep RL. This is a very challenging issue depending on the level of complexity of the observations and the task.

A disentangled representation can be defined as one where we learn independent latent units sensitive to single independent data generative factors \cite{Bengio13}. According to \cite{Bengio13}, the performance of machine learning algorithms can be boosted by taking advantage of a disentangled representation, instead of raw data. In this study, the raw input for SRL, where a robot attempts to grasp a target object in a cluttered environment, is extremely complex. Thus, we hypothesize that the disentanglement of a raw image might help a robot to learn a compact representation that encourages deep RL to attain efficient robotic grasping skills. In this paper, the term \textquote{disentanglement of the raw image} refers to the following condition of the raw image: the raw image is reorganized by separating statistically independent objects or properties. 


To the best of our knowledge, our study is the first to achieve considerable performance in the task of visual grasping of diverse unseen objects via off-policy actor-critic deep RL. In a realistic simulated environment that includes diverse unseen objects, we demonstrate the effectiveness of this approach while varying the levels of disentanglement. 

In summary, our contributions are as follows:

\begin{itemize}
	\item We present three levels of disentanglement for a raw input image. This enables the SRL algorithms to learn a pertinent state representation for the subsequent deep RL. Eventually, robotic grasping skills for diverse unseen objects can be learned based on off-policy actor-critic deep RL.

	\item We investigate effective SRL algorithms given the disentanglement of a raw image. To this end, we first train various SRL algorithms and subsequently perform off-policy actor-critic deep RL using the learned SRL models. The success rate of grasping is used to indirectly evaluate the performances of the SRL algorithms.

\end{itemize}

\section{RELATED WORK}

\textbf{Deep RL for visual grasping.} Most studies on vision-based grasping have applied deep RL methods in limited scenarios, either single objects \cite{Popov17} or simple geometric shapes such as cubes \cite{Zeng18}. Recent approaches presented by Google \cite{Quillen18}, \cite{Kalashnikov18} have successfully extended DQN, which is the most popular value-based RL method, to learn grasping skills for diverse objects via stochastic optimization with massive parallel computational resources. To handle continuous actions, they use a simple stochastic optimization method called the cross-entropy method to compute the \textit{argmax} in the temporal difference (TD) target. However, as mentioned previously, performing this optimization at every decision step is computationally expensive and restricts its widespread use. This is because the stochastic optimization procedure to select an action in these methods require 10--100 times more computation than RL algorithms that explicitly learn policy networks, depending on the number of action samples in the cross-entropy method. 

Meanwhile, off-policy actor-critic deep RL has rarely been applied to visual grasping tasks. A simulated comparative evaluation to investigate which off-policy RL algorithms are best suited for image-based robotic grasping is presented in \cite{Quillen18}. In this study, path consistency learning and DDPG, which explicitly learn an actor network to choose actions, perform poorly compared to the other baselines, which only learn a Q-function and select actions using stochastic optimization.

\textbf{State representation learning in deep RL.} State representation learning (SRL) is a particular case of feature learning in which a set of features to be learned is a vector of a reduced set of the most representative features that are sufficient for efficient policy learning. Examples of generic SRL criteria include reconstructing observations \cite{Finn16}, \cite{Hoof16}, learning the forward or inverse dynamics in a state embedding space \cite{Agrawal16}, \cite{Watter15}, and predicting instantaneous rewards \cite{Jaderberg17}. Some studies considered specific constraints or prior knowledge imposed by physics. Temporal continuity assumes that interesting features fluctuate slowly and continuously through time \cite{Kompella11}, \cite{Jonschkowski15}. Causality generally reflects the principles by which the task-relevant features, together with the action, determine the reward \cite{Jonschkowski15}..

SRL can be jointly performed with the RL \cite{Jaderberg17}, \cite{Zhang19} or be separated from it \cite{Finn16}, \cite{Lee17}. Simultaneous learning of the state-representation and control policy has potential because SRL can aid RL and RL helps shape the state representation. However, realizing this potential can be non-trivial depending on the class of RL and the level of task complexity.

Although the majority of SRL algorithms do not explicitly consider temporal connections among data, the recent Deep Planning Network (PlaNet) \cite{Hafner18} and Stochastic Latent Actor-Critic (SLAC) \cite{Lee19} utilize a latent-space dynamics model, and these approaches achieve a significant sample efficiency for a range of simple image-based control tasks such as the DeepMind control suite \cite{Tassa18}.

\section{Off-Policy Actor-Critic Deep RL Algorithms}

We begin by briefly describing the standard off-policy actor-critic deep RL algorithms used to evaluate the effectiveness of the proposed disentanglement.

\subsection{Deep Deterministic Policy Gradient}

Deep Deterministic Policy Gradient (DDPG) \cite{Lillicrap16a} aims to learn a parameterized deterministic policy $\pi_{\theta}(s)=a$ rather than a probability distribution over all the actions. The objective for training the actor in DDPG is to maximize the expected value of this policy, as follows:

\begin{eqnarray}
\label{eq:actor_loss_DDPG}
J(\theta) = \mathbb{E}_{s,a} [Q_{\pi_{\theta}}(s, a)|_{a=\pi_{\theta}(s)}].
\end{eqnarray}

\noindent Based on the deterministic policy gradient theorem, one can write the gradient of this objective as

\begin{eqnarray}
\label{eq:actor_gradient_DDPG}
\nabla_{\theta}J(\theta) \approx \mathbb{E}_{\rho} [\nabla_{\theta}\pi_{\theta}(s) \nabla_{a}Q_{\pi_{\theta}}(s,a)|_{a=\pi_{\theta}(s)}],
\end{eqnarray}

\noindent where $\rho$ is the state-visitation distribution associated with a certain behavior policy. We can empirically evaluate this gradient using data gathered off-policy and allowing the behavior policy to differ from $\pi$.

The parameterized critic $Q_{\omega}$ is updated by minimizing the TD error, i.e., the difference between the action-value function before and after applying the Bellman update. By taking the mean squared error of the TD error, we can express the resulting loss as

\begin{eqnarray}
\label{eq:critic_loss_DDPG}
\mathcal{L}(\omega) = \mathbb{E}_{\rho} [(r(s,a)+\gamma\mathbb{E}[Q_{\omega^\prime}(s^\prime,\pi_{\theta^\prime}(s^\prime))]-Q_{\omega}(s, a))^{2}],
\end{eqnarray}

\noindent where the TD error will be evaluated under a separate target policy and value networks with separate parameters $(\theta^\prime, \omega^\prime)$ to stabilize learning.

\subsection{Distributed Distributional Deep Deterministic Policy Gradient}

D4PG \cite{Barth18} is an extended version of the DDPG that increases the training and sample efficiency. These extensions include a distributional critic update, the use of distributed parallel actors, N-step returns, and prioritization of the experience replay. Among them, the inclusion of the critic update is the most crucial to the overall performance. Therefore, in this study, we implemented D4PG by only including the distributional critic update, and we additionally adapted the parallel actors to gather grasping experience as quickly as possible.     

To introduce the distributional update, let there be a random variable $Z_{\pi}$, such that $Q_{\pi}(s,a)=\mathbb{E}Z_{\pi}(s,a)$. The parameterized critic $Z_{\omega}$ is updated by minimizing the TD error as in DDPG. We write the loss as

\begin{eqnarray}
\label{eq:critic_loss_D4PG}
\mathcal{L}(\omega) = \mathbb{E}_{\rho} [d(r(s,a)+\gamma\mathbb{E}[Z_{\omega^\prime}(s^\prime,\pi_{\theta^\prime}(s^\prime))]-Z_{\omega}(s, a))],
\end{eqnarray}

\noindent where $d$ is a certain metric that measures the distance between two distributions. Two components that can have a significant impact on the performance of this algorithm are the specific distribution used for $Z_{\omega}$ and the metric $d$. In this study, we used the categorical distribution and Kullback–Leibler(KL) divergence, respectively.

The actor update in this distributional policy gradient algorithm is completed by including the action-value distribution inside the update, which is defined as

\begin{eqnarray}
\label{eq:actor_gradient_D4PG}
\nabla_{\theta}J(\theta) \approx \mathbb{E}_{\rho} [\nabla_{\theta}\pi_{\theta}(s) \mathbb{E}[\nabla_{a}Z_{\pi_{\theta}}(s,a)]|_{a=\pi_{\theta}(s)}].
\end{eqnarray}

\section{State Representation Learning Algorithms}


In this section, we introduce the SRL algorithms used for comparative evaluation in this paper. The objective of these SRL algorithms is to encode some form of prior knowledge that can help simplify and regularize the state representation. In this study, we prefer the model in which the training criteria is based on a more generic prior.

\subsection{Autoencoder-Based Model}

Generally, an autoencoder (AE) is a neural network that is trained to output a close approximation of its input \cite{Vincent10}. An AE consists of an encoder, which maps a high dimensional input such as an image to a reduced vector of latent variables,
and a decoder, which uses this reduced representation to generate a reconstruction of the original high-dimensional input. The AE is optimized by minimizing the following loss: 

\begin{eqnarray}
\label{eq:AE}
\mathcal{L}_{AE}(\theta, \phi) = \mathbb{E}_{p(x)}[ \parallel f_{\phi}(f_{\theta}(x)) = x \parallel^{2}],
\end{eqnarray}

\noindent where $f_{\theta}$ and $f_{\phi}$ denote the encoder and decoder, respectively, and $x$ denotes a high dimensional data point in the training dataset.

Variational AEs (VAEs) \cite{Kingma13} are variations of AEs and are mostly used for a wide range of representation learning. VAEs aim to learn a parametric latent variable model by maximizing the marginal log-likelihood of the training data. By introducing an approximate posterior $q_{\theta}(z|x)$, which is an approximation of the intractable true posterior $p(x|z)$, we can write the loss for VAEs as

\begin{eqnarray}
\label{eq:VAE}
\mathcal{L}_{VAE}(\theta, \phi) = \mathbb{KL}(q_{\theta}(z|x) \parallel p(z))-\mathbb{E}_{q_{\theta}}(z|x)[log p_{\phi}(x|z)],
\end{eqnarray}

\noindent where the second term measures the reconstruction error and the first term is regularization, which quantifies how well $q_{\theta}(z|x)$ matches the prior $p(z)$. Here, $(\theta, \phi)$ represent the parameters of the encoder and decoder model, respectively, which could be neural networks. Higgins et al. \cite{Higgins17} proposed $\beta$-VAE, which weights the second term by a coefficient $\beta > 1$ to encourage it to be factorized into independent factors.

Spatial AEs (SAEs) \cite{Finn16} learn state representation using a variant of the AE objective. SAEs learn a state representation that comprises feature points that encode the image-space positions of distinctive features in the scene. In the encoder, the response maps of the last convolutional layer are first passed through a spatial softmax layer. Then, the expected two-dimensional position (x, y), called the $\textit{feature point}$, for each softmax probability distribution is computed. Hence, the size of this layer is $\textit{the number of channels}$ $\times$ 2.  Finally, the decoder maps the feature points to the input. The SAE is trained by minimizing the reconstruction error, analogous to an AE. However, SAEs place a  particular emphasis on the spatial information of distinctive features using the pair of softmax and expectation operations. Although the complete loss for an SAE includes an additional temporal continuity term that encourages the feature points to change as slowly as possible, we only employ the reconstruction error term.

\subsection{Forward Model}

The general concept of the forward model is to force states to efficiently encode the information required to predict the next state. The model first makes a projection from the observation space to the state space to obtain $s_{t}$ from $x_{t}$ and then predicts $s_{t+1}$ by applying a transition, with the help of $a_{t}$. The loss is computed by comparing the estimated next state $\hat s_{t+1}$ with the value of $s_{t+1}$ derived from the next observation $x_{t+1}$ at the next time step, which is defined as

\begin{eqnarray}
\label{eq:Forward Model}
\mathcal{L}_{FM}(\theta, \omega) = \mathbb{E}_{p(x)}[ \parallel f_{\omega}(f_{\theta}(x_{t}), a_{t}) - f_{\theta}(x_{t+1})| \parallel^{2}],
\end{eqnarray}

\noindent where $s_{t}=f_{\theta}(x_{t})$. Note that the forward models include a mapping from an input $x$ to a state variable $s$; thus, we can combine the observation reconstruction error and the prediction error of the next state. We write the loss as

\begin{eqnarray}
\label{eq:FMAE}
\mathcal{L}_{FM+AE}(\theta, \omega) = \mathcal{L}_{FM}(\theta, \omega)  + \alpha \mathcal{L}_{AE}(\theta, \phi),
\end{eqnarray}

\noindent where the weight $\alpha > 0$ allows us to stress the importance of $\mathcal{L}_{AE}$.

\subsection{Interleaving State Representation Learning and Reinforcement Learning}

In this study, we mainly consider the approach in which the SRL is separated from the RL because the synergistic integration of SRL and RL can be non-trivial. When the auxiliary optimization terms for SRL are added to the RL objective naively, the performance can often decrease rather than increase. However, for the purpose of comparison, we propose a simple method that interleaves SRL and deep RL (i.e., DDPG or D4PG in this paper). This is the most straightforward way of integrating SRL with RL, i.e., simply adding the SRL and RL losses.

\begin{equation}
\begin{array}{r@{}l}
\label{eq:Interleaving}
\mathcal{L}^{Itlv}_{actor} = \mathcal{L}_{actor}  + w_{a} \mathcal{L}_{SRL} \\
\mathcal{L}^{Itlv}_{critic} = \mathcal{L}_{critic}  + w_{c} \mathcal{L}_{SRL},
\end{array}
\end{equation}

\noindent where $w_{a}$ and $w_{c}$ are scaling constants on the SRL loss term and $\mathcal{L}_{actor}$ is the negative of an objective for the actor, e.g., the negative of the objective in Equation~\ref{eq:actor_loss_DDPG}. The three deep neural networks for the actor, critic, and SRL share the convolutional neural network. 

\begin{figure}
 \centering
 \includegraphics[scale=0.6]{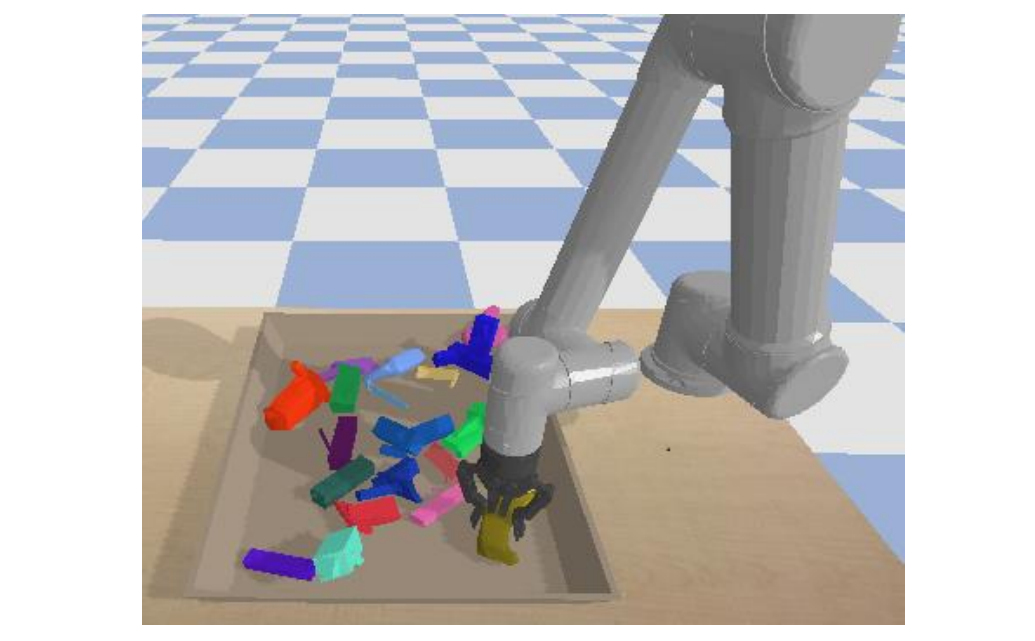} 
 \caption{Illustration of our simulated grasping setup.}
 \label{fig:Simulator}
\end{figure}


\section{Disentanglement of a Raw Input Image}

Most SRL algorithms often are successful when only simple images containing a single object and uncluttered background are provided. They often fail to obtain a pertinent state representation for deep RL given complex images. Because the performance of machine learning algorithms could be improved by taking advantage of a disentangled data, instead of raw data, preprocessing in terms of the disentanglement of a raw input image may be required for SRL in this context. Thus, it is worthwhile examining the eventual effects of different degrees of disentanglement with respect to RL, even if such disentanglements are performed manually. We propose three levels of disentanglement of a raw input image based on various neurophysiological evidence \cite{Roelfsema98}, \cite{Goodale92}. In these studies and succeeding studies investigating the human visual system, the visual attention and separation of $\textit{what}$ and $\textit{where}$ streams in human neural processes were shown to play a significant role in recognition and visually guided behavior. The three levels of disentanglement are as follows:

\begin{figure}
 \centering
 \includegraphics[scale=0.65]{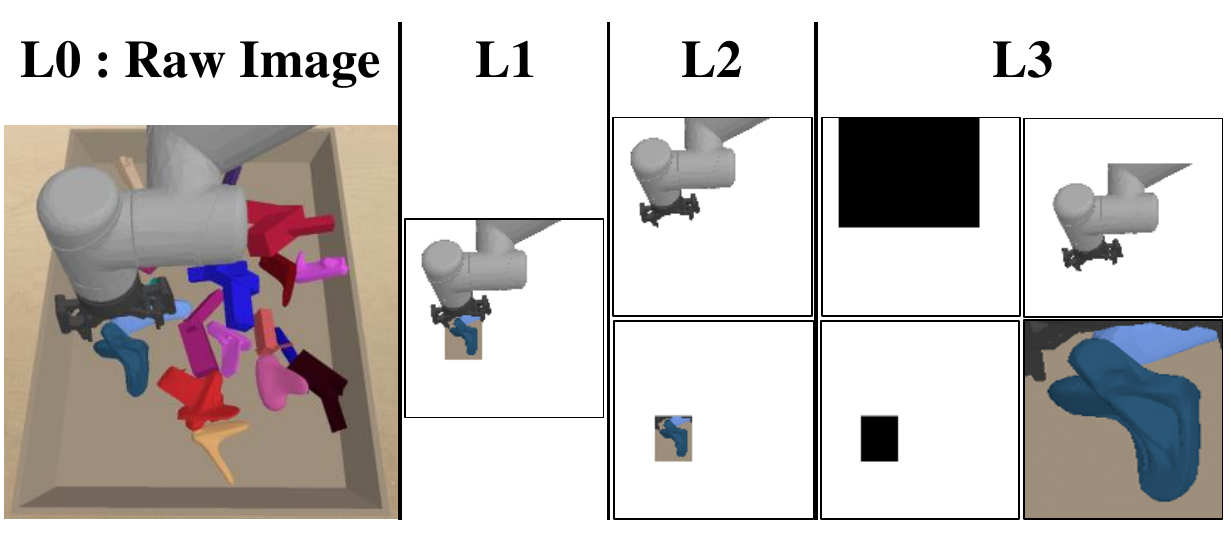} 
 \caption{Example of three levels of disentanglement. At the first level, a robot, a target object, and its periphery are contained in an image. This image is separated into the robot and the focused external world at the second level of disentanglement. There are four images at the third level of disentanglement. Upper Left: position of the robot, Upper Right: visual appearance of the robot, Bottom Left: position of the attended external world, Bottom Right: visual appearance of the attended external world.}
 \label{fig:Decentanglement}
\end{figure}

\begin{itemize}
	\item Visual attention (L1): The regions far from a target object are independent of the target grasping object and robot arm. Hence, we can generate an image in which a robot arm and a fixed-size small region containing the target object are included with the help of standard object detectors \cite{Redmon18}, \cite{Ren15} and semantic segmentation algorithms \cite{Badrinarayanan17}, \cite{He17}. This disentanglement can be crucial for SRL in cluttered scenes. 
	

	\item Separation of internal and external information (L2): For further disentanglement, we separate internal information (i.e., the image that includes only the robot arm and a white background) and external information (i.e., the image that includes a white background and the fixed-size small region containing the target object and its periphery) because they are statistically independent. At this level of preprocessing, the three channels of the RGB image become six channels. 
	
	\item Separation of $\textit{what}$ and $\textit{where}$ streams (L3): The final level of disentanglement divides the $\textit{what}$ and $\textit{where}$ streams for each of the internal and external images. Each image is separated into a position image and a visual appearance image. Those properties are independent of each other. The position image displays the position with a black rectangle tightly covering the robot arm or the focused region. The appearance image is constructed by centering the rectangle tightly covering the robot arm or resizing the cropped region. Thus, the final number of channels after the third level of disentanglement is 12.   
\end{itemize}

\begin{table}
\renewcommand{\arraystretch}{1.1}
\caption{Architecture of the encoder.} \label{tab:encoder}
\begin{center}
\renewcommand{\tabcolsep}{3.0mm}
\begin{tabular}{ccccc}

 & Type & Filter & Size & Output  \\
\hline
 & Convolutional & 16 & 3 x 3 & 128 x 128  \\
 & Convolutional & 32 & 3 x 3/2 & 64 x 64  \\
\hline
\multirow{3}{*}{1x} & Convolutional & 16 & 1 x 1 &   \\
 & Convolutional & 32 & 3 x 3 &  \\
 & Residual &  &  & 64 x 64 \\
\hline         
 & Convolutional & 64 & 3 x 3/2 & 32 x 32  \\
\hline 
\multirow{3}{*}{2x} & Convolutional & 32 & 1 x 1 &   \\
 & Convolutional & 64 & 3 x 3 &  \\
 & Residual &  &  & 32 x 32 \\
\hline         
 & Convolutional & 128 & 3 x 3/2 & 16 x 16  \\
\hline 
\multirow{3}{*}{8x} & Convolutional & 64 & 1 x 1 &   \\
 & Convolutional & 128 & 3 x 3 &  \\
 & Residual &  &  & 16 x 16 \\
\hline         
 & Convolutional & 256 & 3 x 3/2 & 8 x 8  \\
\hline 
\multirow{3}{*}{8x} & Convolutional & 128 & 1 x 1 &   \\
 & Convolutional & 256 & 3 x 3 &  \\
 & Residual &  &  & 8 x 8 \\
\hline         
 & Convolutional & 512 & 3 x 3/2 & 4 x 4  \\
\hline 
\multirow{3}{*}{4x} & Convolutional & 256 & 1 x 1 &   \\
 & Convolutional & 512 & 3 x 3 &  \\
 & Residual &  &  & 4 x 4 \\
\hline         
\multirow{2}{*}{Mean} & Convolutional & 256 & 1 x 1 & \multirow{2}{*}{256}  \\
 & Avgpool & & Global &   \\
\hline  
\multirow{2}{*}{log$\sigma$} & Convolutional & 256 & 1 x 1 &  \multirow{2}{*}{256} \\
 & Avgpool & & Global &   \\
\hline 
\end{tabular}
\end{center}
\end{table}

\begin{table}
\renewcommand{\arraystretch}{1.1}
\caption{Architecture of the decoder.} \label{tab:decoder}
\begin{center}
\renewcommand{\tabcolsep}{3.0mm}
\begin{tabular}{p{0.1mm}ccp{8mm}c}

 & Type & Filter & Size & Output  \\
\hline
 & Deconvolutional & 512 & 3 x 3/8 & 8 x 8  \\
\hline
\multirow{3}{*}{4x} & Convolutional & 256 & 1 x 1 &   \\
 & Convolutional & 512 & 3 x 3 &  \\
 & Residual &  &  & 8 x 8 \\
\hline         
 & Deconvolutional & 256 & 3 x 3/2 & 16 x 16  \\
\hline 
\multirow{3}{*}{8x} & Convolutional & 128 & 1 x 1 &   \\
 & Convolutional & 256 & 3 x 3 &  \\
 & Residual &  &  & 16 x 16 \\
\hline         
 & Deconvolutional & 128 & 3 x 3/2 & 32 x 32  \\
\hline 
\multirow{3}{*}{8x} & Convolutional & 64 & 1 x 1 &   \\
 & Convolutional & 128 & 3 x 3 &  \\
 & Residual &  &  & 32 x 32 \\
\hline         
 & Deconvolutional & 64 & 3 x 3/2 & 64 x 64  \\
\hline 
\multirow{3}{*}{8x} & Convolutional & 32 & 1 x 1 &   \\
 & Convolutional & 64 & 3 x 3 &  \\
 & Residual &  &  & 64 x 64 \\
\hline         
 & Deconvolutional & 32 & 3 x 3/2 & 128 x 128  \\
\hline 
\multirow{3}{*}{4x} & Convolutional & 16 & 1 x 1 &   \\
 & Convolutional & 32 & 3 x 3 &  \\
 & Residual &  &  & 128 x 128 \\
\hline         
 & Deconvolutional & 16 & 3 x 3 & \multirow{3}{*}{128 x 128}  \\
 & Convolutional & same with input & 1 x 1 &   \\
 & hyperbolic tangent & same with input &   &   \\
\hline  
\end{tabular}
\end{center}
\end{table}

\section{Experiments} 

\subsection{Experimental Setup}

Our simulated environment for robotic grasping was constructed in the Bullet simulator \cite{Coumans17}. As shown in Fig.~\ref{fig:Simulator}, the simulated environment uses the configuration of a real UR5 arm, a bin, and a simulated monocular camera. The camera was fixed to capture the robot and the bin in which the objects are placed. Each RGB image is resized and cropped to a resolution of 360 $\times$ 360. We used 200 objects for training and 100 objects for testing. The objects were procedurally generated random geometric shapes published by Google \cite{Google}. The robot’s task was to grasp a randomly selected target object from the bin within a fixed number of timesteps (T = 15). After each episode ends, a new target is selected. The arm moves via the position control of the vertically-oriented gripper.

The performances of the SRL algorithms were indirectly evaluated in this study via the success rate of grasping. In all experiments, five robots operated in parallel, where the training and test phases alternately repeated. In the training phase, 20 objects were randomly chosen from the training dataset for each robot. A robot attempted to grasp the target object, where the maximum number of grasping steps was set to 15. If the robot grasped the target object successfully, the grasped object was removed. When there were no remaining objects in the bin or a collision between the robot and the bin occurred, 20 new objects were selected in the training set. After every 50 grasp attempts, 20 objects were randomly chosen from the test dataset for evaluation. Each robot performed 10 grasping trials. The average of the last success rates produced by each robot was recorded. Most tests continued until the number of training steps reached approximately 300k. In general, the number of grasps was approximately 200k at the end of each test. We mainly used D4PG as the RL algorithm.

\begin{figure}
 \centering
 \includegraphics[scale=0.5]{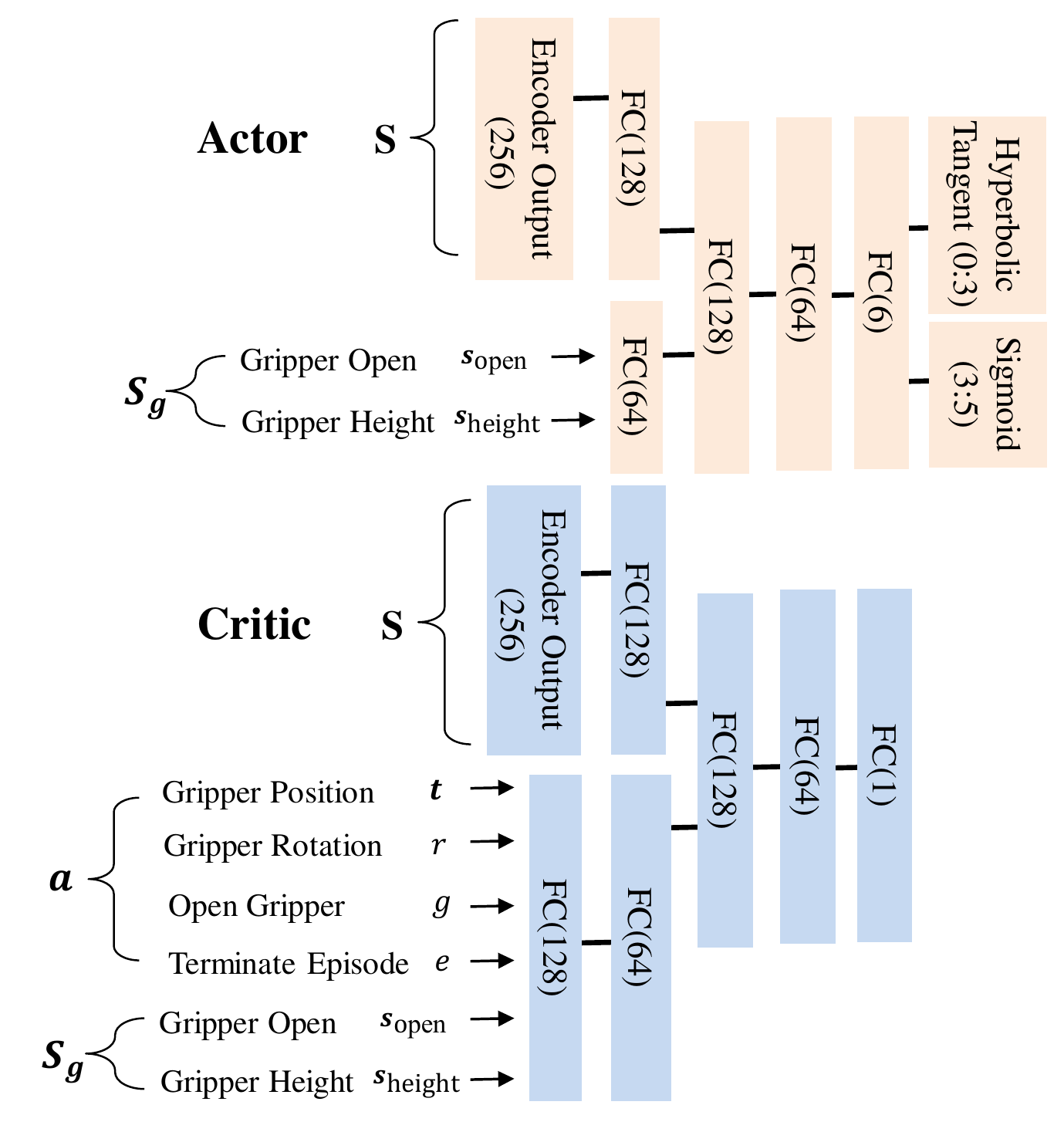} 
 \caption{Neural architecture for the actor and critic.}
 \label{fig:ActorCritic}
\end{figure}

\textbf{Disentanglement of a raw input image.} An example of different levels of disentanglement is shown in Fig.~\ref{fig:Decentanglement}. At L1, the 360 $\times$ 360 image is resized to 128 $\times$ 128. To construct the cropped region containing the target object, the minimum bounding box of the target object is determined and then expanded by 15 pixels in all directions. At L3, the rectangle tightly covering the robot arm is moved to the center of the visual appearance image and the cropped region of the attended external world is resized to an image 128 $\times$ 128 in size.

\textbf{State Representation Algorithms.} We mainly used SAE, $\beta$-VAE ($\beta$=0.1), and FM+AE (i.e., the model in Equation~\ref{eq:FMAE}, which combined the losses of the forward model and AE) to validate effectiveness of the disentanglement. AE, VAE, and FM+$\beta$-VAE were also trained for comparison with various SRL algorithms, given the disentanglement of a raw image. Moreover, we want to check if interleaving SRL and RL can induce a synergistic  performance in our problem setting. Thus, simultaneous learning of $\beta$-VAE and D4PG, referred to as $\beta$-VAE+D4PG, and simultaneous learning of FM+AE and D4PG, referred to as FM+AE+D4PG, were attempted.

\textbf{Markov decision process(MDP) for grasping.} The state $\textbf{s} \in \mathcal{S}$ is provided by an encoder trained via SRL. We also include the current status of the gripper in the state. This additional state comprises a binary indicator of whether the gripper is open or closed and the vertical position of the gripper relative to the floor. The action $\textbf{a} \in \mathcal{A}$ consists of a vector in Cartesian space $\textbf{t} \in \mathcal{R}^{3}$ indicating the desired gripper position, a desired azimuthal angle $r \in \mathcal{R}$, a command for gripper $g \in \mathcal{R}$, and a termination command $e \in \mathcal{R}$ that ends the episode such that $\textbf{a}=(\textbf{t}, r, g, e)$. In practice, the variables for the gripper and termination are considered to be binary when the robot operates. The robot opens the gripper if $g$ exceeds 0.5 and terminates the episode if $e$ exceeds 0.5. The reward is 1 at the end of the episode if the gripper contains a target object and is above a certain height, and it is 0 otherwise. In case of L0 disentanglement, the robot gets reward 1 if the gripper holds an any object and is above the predetermined height. The end of the episode is the timestep at which the model either emits the termination action or exceeds the maximum number of timesteps. The reason for allowing the policy to decide when to terminate itself is that it forces the policy to properly understand the objective of the task. To encourage the robot to grasp more quickly, we provide a small penalty $r(\textbf{s}_{t}, \textbf{a}_{t}) = -0.05$ for all timesteps. The proposed MDP for grasping is very similar to the MDP presented in \cite{Kalashnikov18}. We adopted most of their approach.

\textbf{Structures of Deep Neural Networks.} We used an architecture that is similar to that of Darknet-53 \cite{Redmon18} for the convolutional network shared by the actor, critic, and AE-based model. The full neural network architectures for the encoder and decoder are explained in Tables~\ref{tab:encoder} and ~\ref{tab:decoder}, respectively. The output of the encoder consists of the mean and log of the standard deviation, and the size of each mean and standard deviation vector is set to 256. For the encoder of the SAE, we used four filters with a stride of two for the first convolutional layer and used filters with a stride of one for the other convolutional layers. Thus, the input dimensions for the softmax layer are 64 $\times$ 64 $\times$ 128. Location information should be retained before the expectation layer in the SAE. In the training of AE and RL, only the mean is used. The architectures of the actor and critic are shown in Fig.~\ref{fig:ActorCritic}, which illustrates the neural structures for DDPG. For D4PG, the size of the last fully connected (FC) layer is set to 51 and softmax is subsequently applied because we use categorical distributions. For the forward model, the concatenation of the output of the encoder, status of the gripper, and action is given as the input for the FC layers, whose output size is 256. This was followed by three more FC layers with the same output size. Each convolution and FC layer in all implementations uses batch normalization and ReLU nonlinearity, except for the last convolution and FC layers. The deconvolutional layer uses leaky ReLU.

\textbf{Data collection.} Because an actor has very little success in the early learning stages, we also use a weak scripted exploration policy to obtain more samples from good states and actions. Hence, we used two different exploration policies, $\pi_{scripted}$ and $\pi_{actor}$, instead of just $\pi_{actor}$. Policy $\pi_{scripted}$ performs multi-step exploration in the following manner: An (x, y) coordinate above the center of the target object is chosen, the open gripper is lowered to table level in three random descent steps. The gripper is then closed and it returns to the original height in three ascent steps. Gaussian noise is added to both $\pi_{scripted}$ and $\pi_{actor}$ with zero mean and a standard deviation of 0.05. Initially, $\pi_{scripted}$ is selected with a probability of 1 whereas $\pi_{actor}$ is 0. Then, the probability of $\pi_{scripted}$ decreases linearly until 0.05, being inversely proportional to the current timestep. Policy $\pi_{scripted}$ attains a success rate of approximately 15\% and was originally presented in \cite{Kalashnikov18}. The data collection procedure for the separate learning of SRL was analogous to that in RL, except only $\pi_{scripted}$ was used.  

We were able to generate data with multiple simulators running in parallel and conducted a large scale experiment within a day. All robots shared a replay buffer and stored all of their experiences in it. A learner trained the parameters of the actor and critic using data residing in the replay buffer. Each robot requested a policy for the actor when it acted with $\pi_{actor}$.


\textbf{Parameters.} For RL, we used Adam \cite{Kingma14} to learn the neural network parameters at learning rates of $10^{-4}$ and $10^{-3}$ for the actor and critic, respectively. We used $\gamma$ = 0.99 for the discount factor and $\tau$ = 0.01 for the soft target updates. For all SRL algorithms, we also used Adam, and the learning rate was set to $10^{-4}$. We trained $\beta$-VAE with $\beta=0.1$ and FM+AE with $\alpha=0.1$. In addition, $\alpha$ in Equation~\ref{eq:FMAE} was set to 0.1 and $w_{a}$ and $w_{c}$ in Equation~\ref{eq:Interleaving} were both set to 1.

\begin{figure*}
 \centering
 \includegraphics[scale=0.8]{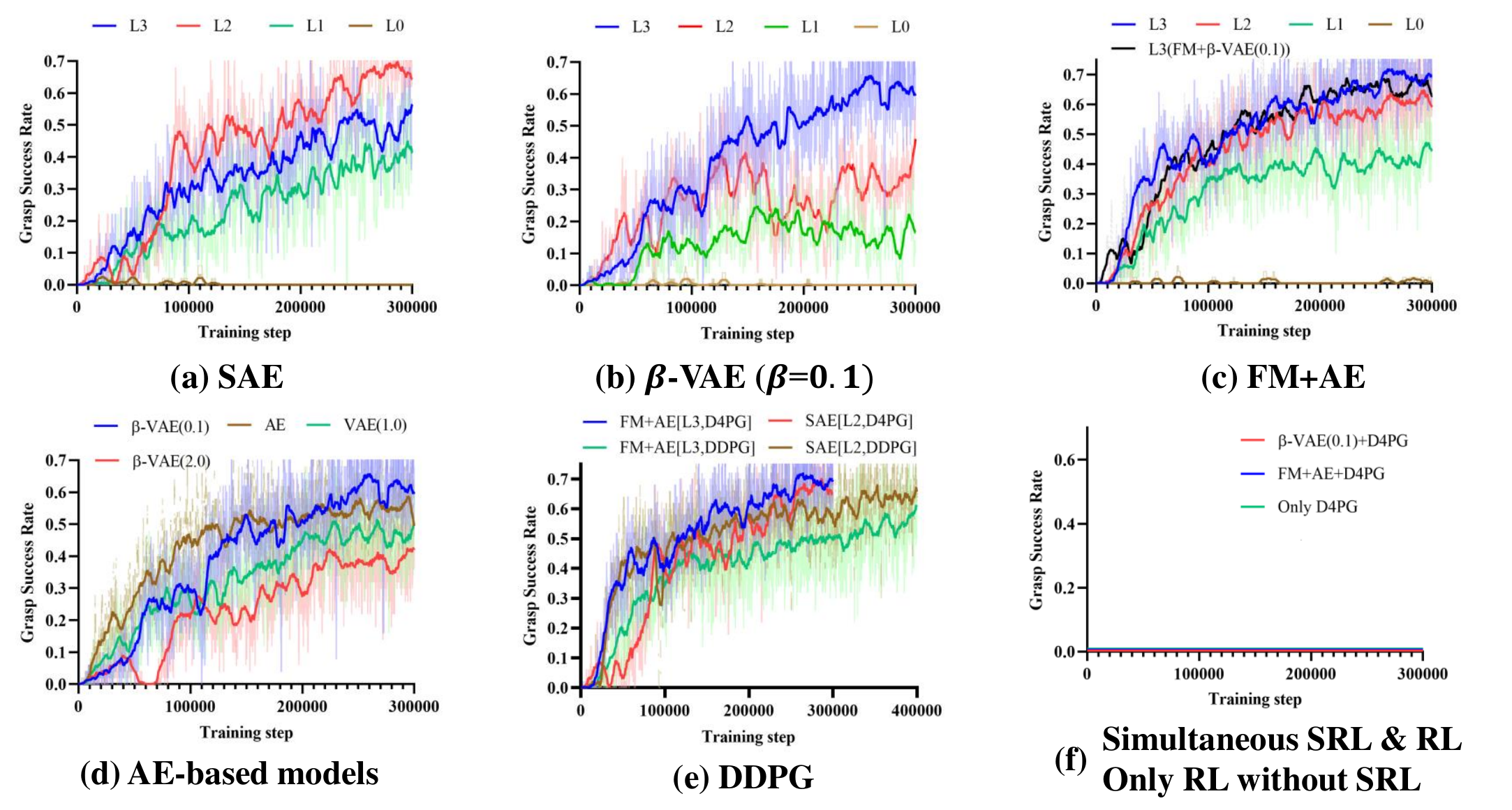} 
 \caption{Mean success rate over the number of training steps. (a)--(c) Comparison of performances given different levels of disentanglement. (d) Comparison of AE-based models. (e) Performances based on DDPG. (e) Grasping success rates of simultaneous SRL and RL methods. (d) and (f) L3 disentanglement used for raw images. The light colored lines and bold lines show actual success rates and temporally averaged results, respectively.}
 \label{fig:Results}
\end{figure*}  
 
\subsection{Experimental Results}




Figure~\ref{fig:Results} shows the mean success rate over the number of training steps, where the light colored lines and bold lines show actual success rates and temporally averaged results, respectively. Because the actual success rate is computed with only five samples, we consider the smoothed lines to evaluate the performance. 

The performance curves for SAE, $\beta$-VAE ($\beta$=0.1), and FM+AE, given different degrees of disentanglement, are shown in Fig.~\ref{fig:Results}(a)--(c), respectively. It is observed that for all SRLs, grasping skills are not learned given raw input images (L0). L1 disentanglement fails to obtain substantial improvement in all three SRL models. Each algorithm in general attains better performances as the level of disentanglement increases. Methods $\beta$-VAE ($\beta$=0.1), and FM+AE achieve their best performances for L3 disentanglement, and SAE performs best at L2 disentanglement. 

The best success rate is 71.9\%, which is achieved by FM+AE. We note that FM+$\beta$-VAE ($\beta$=0.1) given L3 disentanglement achieves a similar performance. It is interesting that SAE attains its best success rate at L2 disentanglement. The SAE does not have a standard encoder structure unlike VAE or FM+AE as the SAE extracts the spatial information of distinctive features in the feature point layer. Presumably the different encoder structure produces the inconsistent results. Overall, these results strongly indicate that the disentanglement of an input image provides significant benefits for various SRL models.  


Figure~\ref{fig:Results}(d) compares the performance curves produced by AE ($\beta$=0), our $\beta$-VAE ($\beta$=0.1), VAE ($\beta$=1), and $\beta$-VAE ($\beta$=2). Our $\beta$-VAE achieves the best results whereas the performances decrease as $\beta$ increases or decreases. A large $\beta$ is generally considered to help a model learn independent data generative factors, such as object identity, position, scale, lighting, or color, which may be essential information for subsequent RL. However, we note that the performance can decrease rather than increase with this hyperparameter setting if a complex image is given as input. 

Figure~\ref{fig:Results}(e) shows the sensitivity of the proposed disentanglement to a different actor-critic deep RL algorithm. For DDPG, the performances curves for SAE and FM+AE are similar or slightly lower than those based on D4PG at the corresponding training steps. However, the final success rates are very close to those of D4PG.  

It is observed in Fig.~\ref{fig:Results}(f) that the simultaneous SRL and RL approach as well as the solely RL-based learning approach completely fail to learn the grasping skill even if L3 disentanglement is given. As mentioned earlier, learning an action-value function and its corresponding actor function jointly is difficult owing to their co-dependencies with respect to each other's output distribution. It seems that including SRL here via the proposed simple integration makes the learning more unstable. In addition, only RL-based learning without a pretrained SRL model also achieves poor performance.   

The best success rate attained in our simulated evaluation is unsurprising compared to that produced by state-of-the-art grasping algorithms \cite{Mahler18}, \cite{Kalashnikov18} in real world settings. However, the real-world studies solved grasping tasks without a specific target object. In our scenario, a target object is selected randomly and surrounding objects often block direct access to the desired object. Moreover, the objects used in our evaluation have highly complex geometric shapes. For a video of this experiment, see https://www.youtube.com/watch?v=LHOpvjsRaZs\&feature=.

\section{Conclusions and Future Works}

To the best of our knowledge, this work is the first to propose a method based on off-policy actor-critic deep RL that achieves a substantial success rate for closed-loop vision-based grasping of diverse unseen objects. Whereas state-of-the-art value-based deep RL requires massive computation to select a proper policy, which makes it impractical, our proposed method is able to perform a successful grasp with requiring affordable computational cost.   

Although the development of a novel representation learning algorithm to accelerate RL is indeed plausible, in this study, we trained SRL models using manually disentangled images. However, we believe that this work is valuable in the sense that it is a first step to investigate the effect of the disentangled raw data on training various SRL models.


A natural extension of this study would be to perform a similar range of tasks in real-world settings. A key remaining challenge to accomplishing this is crossing the ``reality gap,'' which is the difficulty of transferring models trained on simulated images to real world ones. Fortunately, several recent studies have addressed this problem via domain adaptation \cite{Bousmalis18}, \cite{James19} and domain randomization \cite{Tobin17}, \cite{Ren19}. The extension of this work to grasping in real environments by integrating these techniques and SRL based on the disentanglement of a raw input image would be an exciting direction for future work.



\end{document}